\documentclass[10pt,letterpaper]{article}

\usepackage{cogsci}

\cogscifinalcopy 

\usepackage{pslatex}
\usepackage{apacite}
\usepackage{float} 
\usepackage{graphicx}
\usepackage{amsmath}
\usepackage{amssymb}
\usepackage{booktabs}

\usepackage{tabularx}
\newcolumntype{R}{>{\raggedleft\arraybackslash}X}
\usepackage{multirow} 
\usepackage{array}
\usepackage{dcolumn} 
\newcolumntype{d}[1]{D{.}{.}{#1}}

\usepackage[acronym,indexonlyfirst,nomain]{glossaries}
\setacronymstyle{long-short}
\newacronym{SSP}{SSP}{spatial semantic pointer}
\newacronym{SP}{SP}{semantic pointer}
\newacronym{SPA}{SPA}{Semantic Pointer Architecture}
\newacronym{VSA}{VSA}{Vector Symbolic Architecture}
\newacronym{VQA}{VQA}{Visual Question Answering}
\newacronym{NEF}{NEF}{Neural Engineering Framework}
\newacronym{NMN}{NMN}{Neural Module Network}
\newacronym{HRR}{HRR}{Holographic Reduced Representations}

\title{VSA4VQA: Scaling a Vector Symbolic Architecture\\ to Visual Question Answering on Natural Images}

\author{{\large \bf Anna Penzkofer (anna.penzkofer@vis.uni-stuttgart.de)} \\
  University of Stuttgart, Institute for Visualization and Interactive Systems (VIS), Germany
  \AND {\large \bf Lei Shi (lei.shi@vis.uni-stuttgart.de)} \\
  University of Stuttgart, Institute for Visualization and Interactive Systems (VIS), Germany
  \AND {\large \bf Andreas Bulling (andreas.bulling@vis.uni-stuttgart.de)} \\
  University of Stuttgart, Institute for Visualization and Interactive Systems (VIS), Germany}

\begin{document}

\maketitle

\begin{abstract}
While Vector Symbolic Architectures (VSAs) are promising for modelling spatial cognition, their application is currently limited to artificially generated images and simple spatial queries.
We propose \textit{VSA4VQA} -- a novel 4D implementation of VSAs that implements a mental representation of natural images for the challenging task of Visual Question Answering (VQA).
VSA4VQA is the first model to scale a VSA to complex spatial queries.
Our method is based on the Semantic Pointer Architecture (SPA) to encode objects in a hyper-dimensional vector space.
To encode natural images, we extend the SPA to include dimensions for object's width and height in addition to their spatial location. 
To perform spatial queries we further introduce learned spatial query masks and integrate a pre-trained vision-language model for answering attribute-related questions. 
We evaluate our method on the GQA benchmark dataset and show that it can effectively encode natural images, achieving competitive performance to state-of-the-art deep learning methods for zero-shot VQA.

\end{abstract}

\section{Introduction}

\glspl{VSA} have shown significant potential for cognitive modelling \cite{stewart_spaun_2012, komer_neural_2019, lu_representing_2019, bartlett_biologically-based_2022, hersche_neuro-vector-symbolic_2023}.
At the core of VSAs are hyper-dimensional vectors as well as adding and binding operations to build vector compositions that can be reversed and disentangled with almost no loss.
Representing concepts or symbols using these vectors enables VSAs to model cognitive processes with compositionality and systematicity \cite{plate_holographic_2003}.  
While \glspl{VSA} have been used to facilitate abstract reasoning on various tasks, such as Raven's Progressive Matrices \cite{choo_spaun_2018, hersche_neuro-vector-symbolic_2023}, they have also been shown to efficiently encode continuous spaces for building mental image representations \cite{komer_neural_2019, lu_representing_2019} or to improve the robustness in 2D spatial navigation tasks \cite{bartlett_biologically-based_2022}.

\begin{figure}[h!]
  \includegraphics[width=0.91\columnwidth]{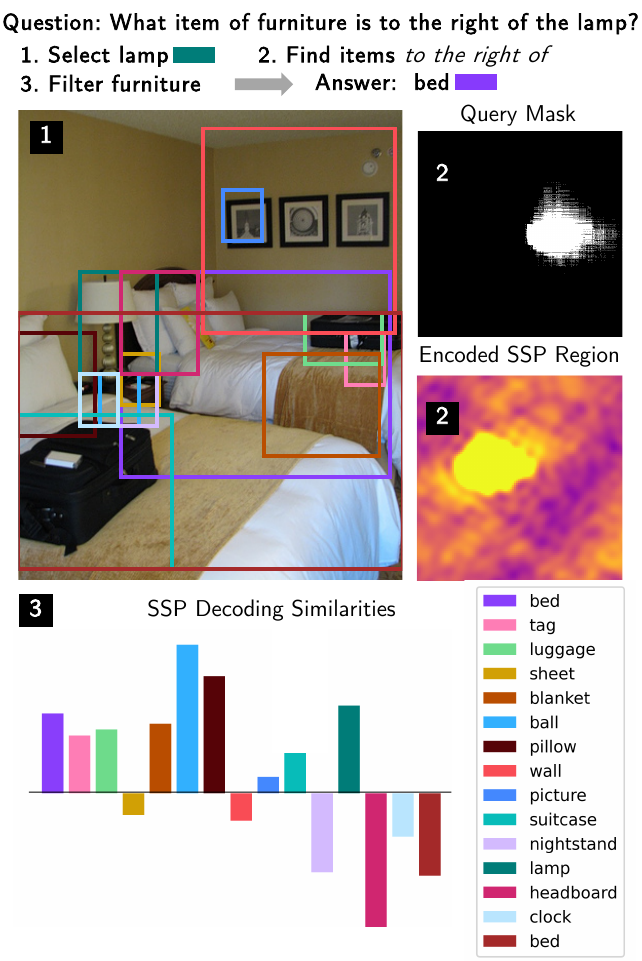}
  \caption{Example question from GQA (Hudson \& Manning, 2019). Our \textit{VSA4VQA} method performs three steps: 1. select the lamp, 2. find items \textit{to the right of} the lamp with a spatial query mask encoded in SSPs, and 3. filter the positive proposals to find furniture, yielding the correct answer ``bed''. }
  \label{fig:example}
\end{figure}

\begin{figure*}[th]
  \centering 
  \includegraphics[width=\linewidth]{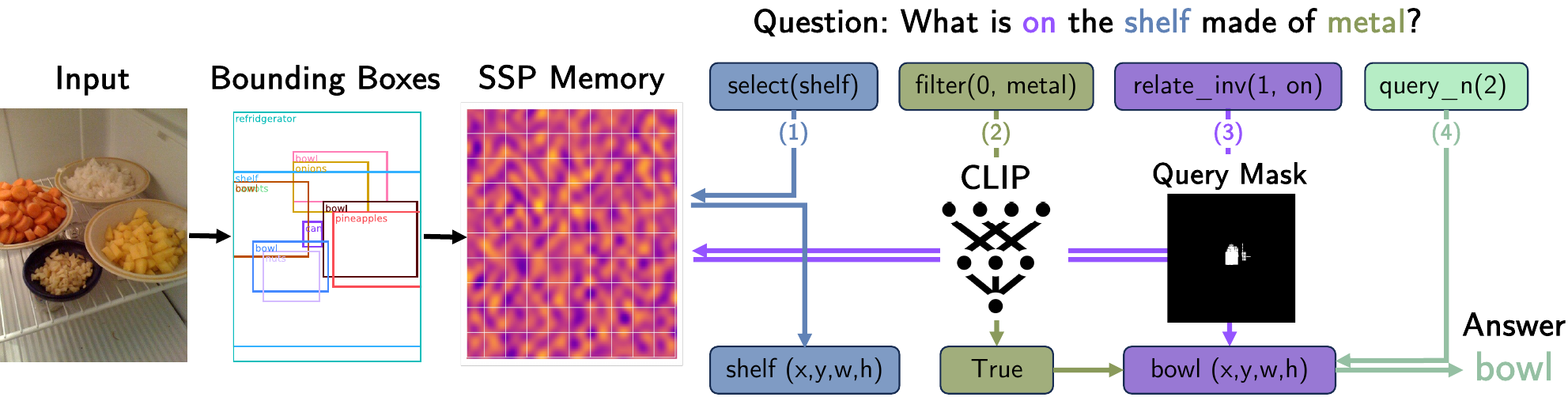}
  \caption{
  Overview of \textit{VSA4VQA}. Object bounding boxes' $(x, y)$-location, width $w$, and height $h$ are encoded into SSP memory.
  The program generator from Chen et al. (2020) maps the question to functions. Our method then implements these functions with (1) SSP unbinding, (2) CLIP, (3) SSP query masks, and finally queries the name of the resulting object (4) to answer the question. For implementation details on all functions see Table~1. 
  }
  \label{fig:pipeline}
\end{figure*}

However, existing VSA models have only been evaluated on artificially generated images with icons \cite{komer_neural_2019} or MNIST digits \cite{lu_representing_2019}, where simple questions about spatial relations between objects were used to evaluate their spatial memory capabilities. 
While underlining the significant potential of VSAs, it is unknown whether these results generalise to images of natural scenes:
Natural images are significantly more visually complex and typically contain numerous objects of different size and appearance.
In addition, natural images exhibit more complex spatial relations between objects that are not covered in artificial images, e.g. objects can be \textit{in front of} or \textit{behind} other objects. 
Answering questions on natural images thus requires a more general understanding of objects' spatial relations \cite{banerjee_weakly_2021}, which is still lacking in current approaches \cite{subramanian_reclip_2022}. 

We propose \textit{VSA4VQA} -- the first VSA model that can answer complex spatial queries on natural images.
VSA4VQA builds on the \gls{SPA} proposed in \cite{eliasmith_how_2013}.
The \gls{SPA} encodes \glspl{SSP} -- hyper-dimensional vectors -- to encode spatial locations in two dimensions.
By encoding objects in an image and binding them to their spatial locations, a cognitively plausible mental representation of the image can be created \cite{komer_neural_2019}. 
Further, the \gls{SPA} allows for querying this mental image representation with spatial regions in relation to other objects.
While previously these spatial queries have only been tested with simple rectangular regions representing the four quadrants of an image, we introduce a novel method to learn query masks that encode spatial relations in natural images \footnote{All query masks and the model code are available at \url{https://perceptualui.org/publications/penzkofer24\_cogsci}}.

To evaluate the mental image representation generated by VSA4VQA, we report experiments on GQA~\cite{hudson_gqa_2019} -- a widely used \gls{VQA} dataset. 
\gls{VQA} is a challenging multimodal task that involves answering questions by reasoning about visual information contained in an image \cite{cao_modularized_2023}.  
VQA is particularly promising as an evaluation task because it requires a compositional understanding of spatial relations~\cite{banerjee_weakly_2021}.
To implement compositional reasoning, our method uses programs generated by a seq-to-seq model~\cite{chen_meta_2020} and assigns each function in the program to a dedicated module~\cite{andreas_neural_2016}.
For functions that query attributes of objects, such as colour or shape, we additionally integrate a pre-trained vision and language model~\cite{radford_learning_2021}.

Taken together, the specific contributions of our work are three-fold: (1) We scale a \gls{VSA} to model cognitively plausible representations of natural images. We further propose zero-shot \gls{VQA} as a particularly suitable task to evaluate the resulting mental image representation in terms of their usefulness for reasoning with complex spatial queries.
(2) We introduce 37 novel spatial query masks that we learn from relation annotations in the GQA dataset and map them to more than 300 spatial relations. (3) We report extensive analyses on error cases and the impact of the dimensionality of \gls{VSA} vectors on the performance.

\section{Method}
Our method performs \gls{VQA} on natural images by implementing a \gls{VSA} as image representation. Figure \ref{fig:pipeline} shows an overview of our method. First, we encode the bounding boxes of all objects in the image into SSP memory. Then, we use programs that map 
questions to functions \cite{chen_meta_2020} to perform sequential reasoning.
To verify attributes that are not present in SSP memory we incorporate CLIP \cite{radford_learning_2021}, a pre-trained vision-language model. For relation queries we use our learned spatial query masks.

\subsection{Image Encoding}
For building a cognitively plausible mental representation of natural images, we extend the fractional binding method of \glspl{SSP} to encode additional dimensions for width and height of objects. To this end, we adjust the mathematical formulation, introduced by \cite{komer_neural_2019}, to include a total number of four dimensions: 
\begin{align}
    \text{SSP} &= \text{SP} \circledast S(x,y, w, h) \\ 
    S(x, y, w, h) &= X^x \circledast Y^y \circledast W^w \circledast H^h \\
    M &= \sum_{i=1}^m \text{SP}_i \circledast S(x_i,y_i, w_i, h_i)
\end{align}
Here, a \gls{SP} is a hyper-dimensional vector with fixed dimension, which is a compressed representation of an object in the image. $x$ and $y$ are the object's spatial locations in the coordinate system of the image, while $w$ and $h$ encode width and height of the object's bounding box. Binding all \glspl{SP} representing an object with the object's location, width, and height and summing them, yields the mental image representation $M$, further referred to as \gls{SSP} memory.

Before encoding images, we set four random \gls{SP} vectors as (X, Y, W, H)-axes and pre-compute all location vectors in the \gls{SSP} vector space.
This yields a discretised grid of 100$\times$100$\times$10$\times$10 points and constitutes our clean-up memory (see \cite{lu_representing_2019} for details). 
The images in GQA vary in size and orientation, therefore, the longer side is always selected to be scaled to fit within the 100$\times$100 vector space. 
For building the \gls{SSP} memory $M$, we need bounding boxes for each object in the image. For this, we use the ground truth object annotations given in the scene graphs of the GQA dataset.

For each detected object we generate a random \gls{SP}.
We then compute the objects point $S(x,y,w,h)$ as defined in Equation 2 and bind the object's \gls{SP} and point $S$ together as described in Equation 1. We add each object to the image's \gls{SSP} memory using element-wise addition, which is the superposition operation in \gls{HRR}.

\begin{table*}[t]
    \caption{Program function types and our implementation. Functions within one type are sorted by the number of arguments. For details on function types $\texttt{exist}$, $\texttt{common}$, $\texttt{different}$, $\texttt{same}$, $\texttt{and}$, $\texttt{or}$ see the original implementation by Chen et al. (2020).}
    \label{tab:funcs}
    \centering
    \small
    \begin{tabularx}{\textwidth}{lp{4.2cm}Xcll} 
         \toprule
            \textbf{Type} & \textbf{Function} &  \textbf{Implementation} &   \textbf{Args} & \textbf{Output} & \textbf{Example}\\
         \midrule
        Select & $\texttt{select}$ & SSP Unbinding & 1 & Position & $\texttt{select(tool)}$ \\[.3em]  
        Relate & $\texttt{relate}$, $\texttt{relate\_inv}$ & SSP Query Mask & 2 & Proposal & $\texttt{relate(basket, on)}$\\ 
        & $\texttt{relate\_name}$, $\texttt{relate\_inv\_name}$  & SSP Query Mask & 3 & Proposal & $\texttt{relate\_name(man, with, hat)}$\\[.3em]
        
        Filter & $\texttt{filter\_v}$, $\texttt{filter\_h}$ & Position & 2 & True / False & $\texttt{filter\_h(bottle, left)}$\\ 
        & $\texttt{filter}$, $\texttt{filter\_not}$ & CLIP & 2 & True / False & $\texttt{filter(refridgerator, red)}$\\[.3em] 
        
        Verify& $\texttt{verify\_f}$ & CLIP & 1 & True / False & $\texttt{verify\_f(beach)}$ \\
        & $\texttt{verify}$ & CLIP & 2 & True / False & $\texttt{verify(umbrella, black)}$ \\ 
        & $\texttt{verify\_rel}$, $\texttt{verify\_rel\_inv}$ & SSP Query Mask & 3 & True / False & $\texttt{verify\_rel(fry, on, tray)}$ \\[.3em]  
        
        Choose & $\texttt{choose\_v}$, $\texttt{choose\_h}$ & Position & 1 &Answer & $\texttt{choose\_v(car, top, bottom)}$\\
        & $\texttt{choose\_f}$ & CLIP & 2 & Answer & $\texttt{choose\_f(indoors, outdoors)}$\\ 
        & $\texttt{choose\_subj}$ & CLIP & 3 & Answer & $\texttt{choose\_subj(boy, man, older)}$\\ 
        & $\texttt{choose\_attr}$ & CLIP & 4 & Answer & $\texttt{choose\_attr(car, color, gray, red)}$ \\ 
        & $\texttt{choose\_rel\_inv}$ & SSP Query Mask & 4 & Answer & $\texttt{choose\_rel(boy, car, left, right)}$ \\[.3em]

        Query & $\texttt{query\_n}$, $\texttt{query\_v}$, $\texttt{query\_h}$ & Position & 1 & Answer & $\texttt{query\_h(chair)}$ \\ 
        & $\texttt{query\_f}$ & CLIP & 1 & Answer & $\texttt{query\_f(place)}$ \\ 
        & $\texttt{query}$ & CLIP & 2 & Answer & $\texttt{query(fence, material)}$\\ 
        \bottomrule 
    \end{tabularx}
\end{table*}

\subsection{Query Masks}
In previous work \cite{komer_neural_2019, lu_representing_2019} it was shown that regions can be encoded as a sum of spatial locations. Such encoded regions were then used to query all objects located within the respective area. However, these region queries were limited to simple rectangles encoding the four quadrants of an image. In contrast, we learn representations for 37 unique query masks, which enables the encoding of more than 300 spatial relations.

The query masks are generated from the relation annotations available in the scene graphs of the training split of GQA \cite{hudson_gqa_2019}. 
The dataset provides $50.65$ relations per image on average, while samples per relation range from two (\textit{shorter than} / \textit{bigger than}) to over one million
(\textit{to the right of} / \textit{to the left of}). For each relation with more than 1,000 samples we generate a spatial query mask by adding up all samples of relative objects, i.e. the objects that the relation applies to.
However, before adding each sample, all objects need to be normalised: 
(1) both objects bounding boxes are scaled so that the bounding box of all queried objects is uniform (50$\times$50 pixels) and (2) the new position of this bounding box is extracted and used to calculate the translation vector for moving it into the centre of a 500$\times$500px image. 

\begin{figure}[th]
  \includegraphics[width=\columnwidth]{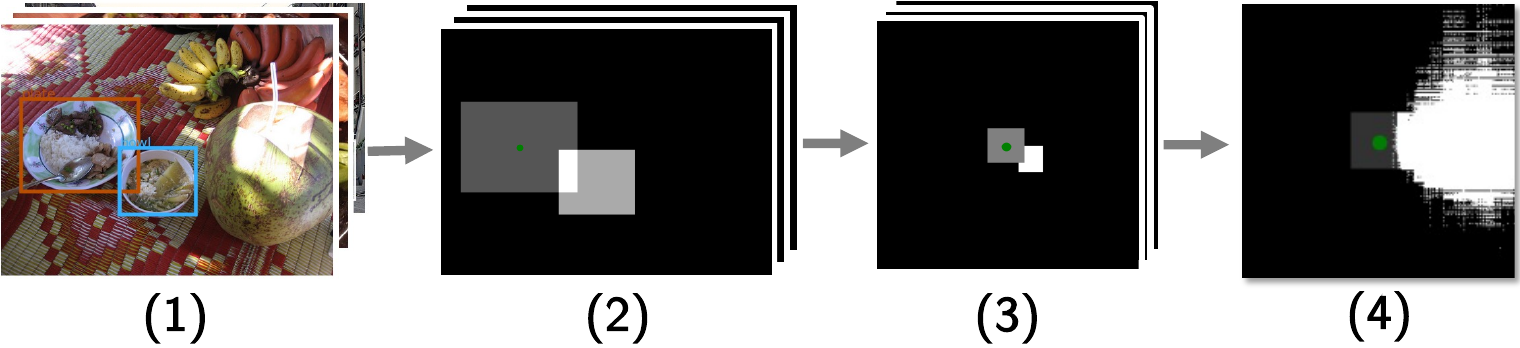}
  \caption{Query mask generation for relation \textit{to the right of}. For all samples of the relation (1): create object masks (2), normalise masks to same scale and center position (3), and add them to obtain final relation mask (4).}
  \label{fig:query}
\end{figure}

In Figure \ref{fig:query}, we show the full process of the query mask generation for relation \textit{to the right of}. 
After normalisation, the areas of all objects that are in relation to the queried object are summed up and averaged. Thresholding the resulting image to exclude areas that have only appeared in less than 5\% of samples, yields the binary query masks depicted in the right-most image. 
The rest of the 37 query masks are obtained in the same way and yield qualitatively plausible representations of the respective relations.  

\subsection{Programs \& Modules}
To answer natural language questions on images, we follow the \gls{NMN} paradigm \cite{andreas_neural_2016}, where questions are mapped to programs that can be executed in sequential steps by separate modules. 
Similar to Mod-Zero-VQA \cite{cao_modularized_2023}, the state-of-the-art deep learning approach for zero-shot \gls{VQA}, we use the programs generated by the pre-trained sequence-to-sequence model of Chen et al. \citeyear{chen_meta_2020}. For the GQA dataset \cite{hudson_gqa_2019}, the programs consist of 10 different function types (see Table \ref{tab:funcs}) and 48 fine-grained functions, which are chained together yielding a list of steps to be performed sequentially. For each step, the result of the respective function is saved for access in the following steps. Almost all programs start with the selection of an object, only exceptions are ``full'' functions that query the entire scene, which are denoted with an \texttt{\_f}. The lengths of programs are up to 9 steps. While the functions $\texttt{relate}$, $\texttt{filter}$, and $\texttt{verify}$ produce outputs that can be reused in following steps, $\texttt{choose}$ and $\texttt{query}$, directly yield answers and terminate the program. 

For example in Figure \ref{fig:pipeline}, the question ``What is on the shelf made of metal?'' is mapped into four functions: \texttt{select(shelf)}, \texttt{filter(0, metal)}, \texttt{relate\_inv(1, on)}, and \texttt{query\_name(2)}. 
The result of the selection from SSP memory is reused in step two, where it is filtered to be made from metal. If indeed it is a metal shelf, the program continues with step three, where all objects are queried that are on the shelf. This returns the SSP of a bowl as the most likely candidate, which is returned in the final step.

We implement five of the program functions with our \gls{SSP} memory. With the image encoded in the \gls{SSP} memory, we can use the unbinding operation $M \circledast OBJ^{-1}$ to perform the $\texttt{select}$ function for any encoded object. 
In general, the unbinding operation yields the respective \gls{SSP} and the object's encoded $(x, y, w, h)$-coordinates.
For performing relation queries, our generated query masks are loaded and encoded as a region in \glspl{SSP}, then, shifted to the correct position. The region query returns all objects that are positively similar to the queried region as proposals. If the function does not specify the relative object it is looking for, the proposal with the highest similarity is returned. If there is an additional attribute in the function that specifies the name of the queried object, the proposal that matches the name or the most similar object that belongs to the named class is returned. 

\subsection{CLIP Integration}
For functions that require attribute verification, we employ the pre-trained model CLIP \cite{radford_learning_2021}.
CLIP is a large vision-language model, trained on 400 million image-text pairs without labels. The model learns visual representations with natural language supervision by enforcing a joined embedding of texts and images. With this, CLIP is capable of predicting which text has the highest similarity to an image and can be used for zero-shot image classification. 
Recently, Shtedritski et al. \citeyear{shtedritski_what_2023} showed that visual prompt engineering CLIP by drawing a red circle around a part of the image, where objects of interest are located, improves performance.
This method enables the use of location information gained from our \gls{SSP} memory, while still offering CLIP an image with global information to provide context. 

We use CLIP to $\texttt{filter}$, $\texttt{verify}$, $\texttt{choose}$, and $\texttt{query}$ attributes of objects. In detail, we use an attribute dictionary and generate sentences based on the attribute type and the object of interest, e.g. ``The colour of the chair is red''. We feed these sentences to CLIP with the corresponding image, where a red circle marks the object of interest. CLIP then returns the likelihood for each image sentence pair. By selecting the sentence with the highest similarity, we find the most likely attribute. Attribute types include colour and shape, but also more difficult types such as weather, materials, or age.    

\section{Experimental Design}
We evaluate the overall performance of our \gls{SSP} representation of natural images and the corresponding reasoning capabilities by computing the zero-shot accuracy on the GQA dataset \cite{hudson_gqa_2019}. \gls{VQA} is a difficult task that requires object detection, scene understanding, and spatial relation understanding. Zero-shot further defines that no training on the specific data set occurs \cite{cao_modularized_2023}, ensuring the generalisability of the methods used. 
We have selected the GQA dataset for evaluation, as it specifically focuses on spatial reasoning \cite{banerjee_weakly_2021}. We evaluate our method on the GQA validation set, which consists of 132,062 questions, paired with 10,234 unique images. Around 54\% of questions in the validation set include a spatial relation query. We hypothesise that the fractional binding method of \glspl{SSP} excels at such spatial queries and specifically analyse this by evaluating the performance of our method on the different types of questions.
Further, we analyse the types of errors our reasoning pipeline exhibits. 

Scaling \glspl{VSA} to \gls{VQA} on natural images requires extensions to previous \gls{SSP} methods \cite{komer_neural_2019, lu_representing_2019}. For instance, we extended the 2D \glspl{SSP} to four dimensions. As a result, the capacity of our \gls{SSP} memory needs to be higher than in the 2D case, where 512 dimensions were found to perform best, while being in line with human capacity limits of working memory \cite{lu_representing_2019}. We analyse the decoding capacity of our 4D \gls{SSP} memory for dimensionalities: 512, 1,024, and 2,048.
Specifically, we compute the mean squared error (MSE) of recalled 2D locations, the intersection-over-union (IoU) of the decoded objects bounding boxes, and the percentage of correctly recalled objects in one image, where correct is defined as $IoU > 0.5$. IoU is calculated as the overlapped area divided by the area of union of two bounding boxes: $\text{IoU} = \frac{box_1 \cap box_2}{box_1 \cup box_2}$ and is a common metric that evaluates the accuracy of bounding boxes.  

Another novelty of our method is the generated query masks for spatial relation queries in natural language. We evaluate our 37 generated query masks qualitatively by comparing them to the common understanding of the spatial relation. For example, the relation \textit{to the right of}, which is depicted in the query mask in Figure \ref{fig:example}, shows the region of interest to be on the right side of the queried object that is set in the centre of the mask.

\section{Results}
\begin{figure}[t]
    \centering
  \includegraphics[width=\columnwidth]{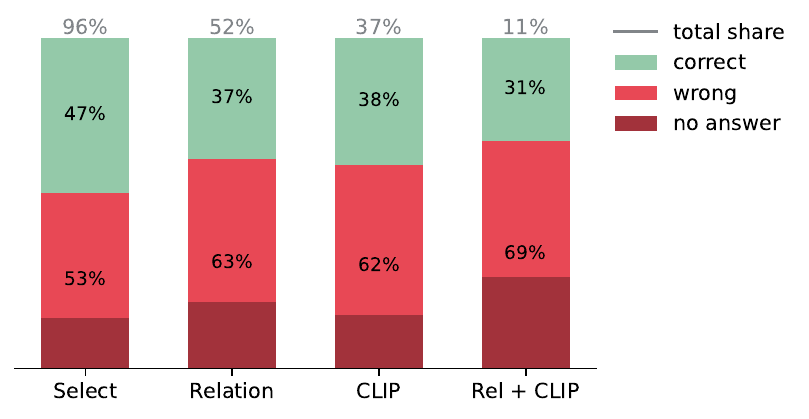}
  \caption{Error analysis by type of question. Percentage on top indicates total share of questions that include the respective query in their program. Wrong answers are further split to indicate when no answer was given.}
  \label{fig:error}
\end{figure}

 Following \cite{cao_modularized_2023, song_clip_2022}, we calculate the overall accuracy of our model \textit{VSA4VQA} on the GQA validation set with three different random seeds and achieve 46.5\% (std 0.09\%). See Table \ref{tab:results} for comparison to other zero-shot methods. The current state-of-the-art deep learning method Mod-Zero-VQA \cite{cao_modularized_2023} also employs the \gls{NMN} paradigm \cite{andreas_neural_2016} and 
uses three different pre-trained deep learning models to implement the functions for answering the questions. However, their method does not encode a cognitively plausible image representation and requires 621 million model parameters, i.e.significantly more than the 150 million used by CLIP.

In Figure \ref{fig:error}, we show a more detailed performance analysis split across the different modules. Questions that use a relation function, implemented via query masks on \gls{SSP} memory, in comparison to attribute functions that are implemented with CLIP. While there are more questions with relational queries (52\%), accuracy of CLIP and relations implemented via \gls{SSP} memory are almost the same, i.e. 37\% in comparison to 38\%. 
However, it is important to note here that the percentages given are based on the overall correct answer to the question, so they involve additional steps, where errors could occur. For example, 96\% of questions include a \texttt{select} function and while the \gls{SSP} unbinding operation correctly decodes $87$\% of items, the accuracy of questions including the select operation is also only 47\%. The number of questions, where both relational queries and attribute queries were combined are shown in the right bar, where accuracy further decreased as more steps are needed. 

A possible reason for the wrong answers in relational queries is the size of the query masks, where it was suggested \cite{lu_representing_2019} that a higher region size will reduce decoding accuracy. We have tested this hypothesis by calculating the pearson correlation coefficient between accuracy and query mask size, resulting in a weak correlation of $-0.30$ (one-sided $p = 0.09$). 
Another possible source of errors is the inherent stochasticity in the \gls{SPA}. Each \gls{SP} is chosen randomly, therefore, the vectors are sometimes more or less suitable for the question. We have balanced this effect by running multiple random seeds, however, running the same question with multiple random seeds and choosing the majority answer might further improve performance. This was computationally infeasible in our case, due to the large dataset size, but should be tested in future work on other datasets. 

\begin{table}[t]
    \begin{center} 
        \caption{Overall accuracy of zero-shot models on GQA validation set. Our method VSA4VQA is comparable to state-of-the-art deep learning method Mod-Zero-VQA.} 
        \label{tab:results} 
        \setlength\extrarowheight{3pt}
        \begin{tabular}{lc}
            \hline
            \textbf{Method}   &  \textbf{Accuracy} \\
            \hline
            TAP-C {\footnotesize \cite{song_clip_2022}}  & $36.3$\% \\
            Mod-Zero-VQA {\footnotesize \cite{cao_modularized_2023}} & $47.3$\% \\
            \hline 
            \hline
            VSA4VQA {\footnotesize (ours)} & $46.5\%$ \\
            \hline
        \end{tabular} 
    \end{center} 
\end{table}

We have further highlighted the amount of questions with no answer in Figure \ref{fig:error}. These occur if a \texttt{select} operation did not find the requested object or if a \texttt{filter} operation returns the wrong result and incorrectly terminates the program. Other causes also include functions that are not implemented due to nesting, e.g. finding objects that are the same or different.  In relation queries this amounts to 20\% and in CLIP queries to 16\%. We highlight these questions as open directions, where further improvements are possible. 

\subsection{Capacity Analysis}

To analyse the impact of model capacity on performance, reflected in the dimensionality of the SSP vectors, we evaluated on a reduced set of 10,000 questions to save computation time, as done in previous work \cite{komer_neural_2019}.
Performance results for SSP vector dimensions of size 512, 1,024, and 2,048 are summarised in Table \ref{tab:capacity}.
512 dimensions were chosen in previous work as it achieved high decoding accuracy for up to 10000 random \glspl{SSP} \cite{komer_neural_2019}. 
As can be seen from the table, here, 512 dimensions result in a high MSE and low IoU between bounding boxes, which is most likely due to the increase in dimensions of the vector space from 2D to 4D.  
At the same time, however, the choice of dimensions does not seem to significantly impact the accuracy: Using 512-dimensional vectors results in similar performance as for higher dimensions. 
This suggests that \gls{SSP} decoding accuracy only plays a minor role in overall performance of \textit{VSA4VQA} on question answering. Further, 2,048 dimensions do not significantly increase accuracy, which indicates that 1,024 dimensions are sufficient, while requiring less computation time and resources. 

\begin{table}[htbp]
    \begin{center} 
        \caption{Capacity analysis. Decoding accuracy of items with different dimension of \gls{SSP} vectors, measured with mean-squared-error (MSE), intersection-over-union (IoU), percentage of correct items, and VQA accuracy.} 
        \label{tab:capacity} 
        \setlength\extrarowheight{3pt}
        \begin{tabular}{ccccc} 
            \hline
            \textbf{Dimensions} &  \textbf{MSE} $\downarrow$ & \textbf{IoU} $\uparrow$ & \textbf{Items} $\uparrow$ & \textbf{Accuracy} $\uparrow$ \\
            \hline
            512 & $51.46$ & $0.59$ & $63.47$\% & $45.03$\%\\
            1,024 & $26.23$ & $0.80$ & $86.71$\% & \textbf{46.02\%}\\ 
            2,048 & \textbf{23.09} & \textbf{0.84} & \textbf{89.78\%} & $45.96$\%\\
            \hline 
        \end{tabular} 
    \end{center} 
\end{table}

\subsection{Error Analysis}
We further analyse the types of errors that can occur on a few selected questions. We found four distinctive types of errors: (1) wrong SSP encoding/decoding, (2) wrong CLIP prediction, (3) bad programs, and (4) ambiguous questions. In the first case, the \texttt{select} method returns the wrong location for the requested object. As we have seen in the capacity analysis, this might be due to the vector's dimension and subsequent noise. When there are many objects, the orthogonality principle of the hyper-dimensional vector space no longer holds and object \glspl{SSP} are no longer orthogonal in the vector space, which in turn results in overlapping \glspl{SSP} that cannot be disentangled correctly in the clean-up process. 

In the second case, wrong CLIP predictions can lead to a false answer or no answer at all. CLIP is sensible to the selection of proposal sentences. We have tested different options: only giving the attribute, using the object name plus the attribute, and building a full sentence.
We found that the full sentence, where possible, gives the best result  
On the other hand, CLIP is also sensible to the processing of the image. We have selected the red circle method proposed by Shtedritski et al. \citeyear{shtedritski_what_2023}, as it showed improved performance. 

The third type of error is due to incorrect programs. Here, we found two main causes: terms with more than one word and incorrect split of \texttt{filter\_v} and \texttt{filter\_h}. While the latter could be remedied in our implementation, the former is difficult to address. For example, the term ``soccer ball'' gets incorrectly split so that the program looks like \texttt{select(soccer)}, which naturally does not yield a result when querying the \gls{SSP} memory, where only \texttt{soccer ball} was encoded. A possible solution would be to have a dictionary with all terms that encompass more than two words and recreate the programs with a respective check of the dictionary. This will be left for future work. 

The final type of error stems from ambiguous questions in the GQA dataset. For example, one question is ``who is wearing a shirt?'', but in the corresponding image both a girl and a boy are wearing shirts, while the correct answer is ``girl.'' This example illustrates that questions can be ambiguous, i.e. even humans might answer them incorrectly. In fact, human subjects tested on 4000 random questions from GQA only achieved an average accuracy of 89.3\% \cite{hudson_gqa_2019}. Approaches on other datasets have addressed this issue by using soft VQA scores that account for multiple correct answers, for details see \cite{cao_modularized_2023, song_clip_2022}. GQA, however, does not provide the required annotations.

\section{Discussion}
In our detailed error analysis we found four causes for errors: wrong \gls{SSP} decoding, incorrect CLIP predictions, bad programs, and ambiguous questions. While the latter two are due to the dataset selection and might not be as pronounced on different datasets, the first two could be improved upon. Specifically, the incorrect CLIP predictions might be circumvented by extracting the attributes such as colour or shape with more specialised feature extractors and encoding them into the \gls{SSP} memory. However, such a feature extractor model needs to be trained and as we have seen in our capacity analysis, the additional encoding of attributes in \glspl{SSP} would require higher vector dimensions. 

We have chosen to use 1,024 dimensional \gls{SSP} vectors to stay comparable to previous work and within human capacity limits \cite{komer_neural_2019}. For future work, however, it would make sense to distinguish between a real memory task, where all objects need to be retained in memory over time, and the \gls{VQA} task, where the image stays available. The former setting is subject to biological capacity limits and should therefore be implemented with limitation on vector size. \gls{VQA}, on the other hand, could allow for higher vector dimensions without loss of biological realism.   

A current limitation of our method is the use of ground truth object annotations. In future work, we want to address this by using pre-trained object detectors.
While GQA provides detections of a fine-tuned Faster-RCNN \cite{ren_faster_2015}, they do not include the labels of bounding boxes, which are critical for our approach and could not be reproduced.
Furthermore, handling multiple objects of the same type is difficult in the current method. In previous work \cite{komer_neural_2019} this was solved by computing an average across multiple random seeds.

Similarly, we found that encoding objects as regions did not work in our setting. This is, again, likely due to the limitation of random seeds -- this approach would allow for decoding the full region with higher precision \cite{lu_representing_2019} but is not possible for the computationally more demanding \gls{VQA} task on natural images. Instead, we chose to extend the \gls{VSA} to four dimensions to encode objects' width and height. In future work, we are interested in finding a more cognitively inspired method. In general, \textit{VSA4VQA} could be improved by moving towards a fully neural implementation on dedicated hardware, which would improve efficiency and, therefore, allow for probabilistic inference with multiple random seeds.    

\section{Conclusion}
Our proposed model, \textit{VSA4VQA}, is capable of answering complex compositional questions on natural images. We have tested our model on a dataset that focuses on spatial queries and achieved comparable performance to current zero-shot deep learning approaches. To the best of our knowledge, \textit{VSA4VQA} is the first model to implement a cognitively plausible image representation of natural images, which can be used to answer complex spatial queries. We have effectively scaled a VSA to encode additional dimensions for width and height of objects. Further, we generated 37 spatial query masks from data to answer relation-based questions and integrated a pre-trained vision-language model to answer attribute-related questions. Our extensive analysis on questions, errors, and capacity limits provides valuable insights for future work.

\section{Acknowledgments}
Anna Penzkofer and Lei Shi were funded by the Deutsche Forschungsgemeinschaft (DFG, German Research Foundation) under Germany’s Excellence Strategy -- EXC 2075 -- 390740016.

\bibliographystyle{apacite}

\setlength{\bibleftmargin}{.125in}
\setlength{\bibindent}{-\bibleftmargin}

\bibliography{references}

\end{document}